\documentclass{article}

\usepackage{PRIMEarxiv}

\usepackage[utf8]{inputenc} % allow utf-8 input
\usepackage[T1]{fontenc}    % use 8-bit T1 fonts
\usepackage{hyperref}       % hyperlinks
\usepackage{url}            % simple URL typesetting
\usepackage{booktabs}       % professional-quality tables
\usepackage{amsfonts}       % blackboard math symbols
\usepackage{nicefrac}       % compact symbols for 1/2, etc.
\usepackage{microtype}      % microtypography
\usepackage{lipsum}
\usepackage{fancyhdr}       % header
\usepackage{graphicx}       % graphics
\graphicspath{{media/}}     % organize your images and other figures under media/ folder

\usepackage{makecell}

\title{Analysing similarities between legal court documents using natural language processing approaches based on Transformers}

\author{\textbf{Raphael Souza de Oliveira} \\
  TRT5 - Regional Labour Court of the 5th Region, Salvador, BA, Brazil \\
  Stricto Sensu Department, SENAI CIMATEC University Center, Salvador, BA, Brazil \\
  \texttt{raphael.oliveira@gmail.com} \\
  \textbf{Erick Giovani Sperandio Nascimento} \\
  Surrey Institute for People-Centred AI, School of Computer Science and Electronic Engineering, \\Faculty of Engineering and Physical Sciences, University of Surrey, Guildford, UK \\
  Stricto Sensu Department, SENAI CIMATEC University Center, Salvador, BA, Brazil \\
  \texttt{erick.sperandio@surrey.ac.uk; erick.sperandio@fieb.org.br}
}

\begin{document}
\maketitle

\begin{abstract}
Recent advances in Artificial Intelligence (AI) have leveraged promising results in solving complex problems in the area of Natural Language Processing (NLP), being an important tool to help in the expeditious resolution of judicial proceedings in the legal area. In this context, this work targets the problem of detecting the degree of similarity between judicial documents that can be achieved in the inference group, by applying six NLP techniques based on the transformers architecture to a case study of legal proceedings in the Brazilian judicial system. The NLP transformer-based models, namely BERT, GPT-2 and RoBERTa, were pre-trained using a general purpose corpora of the Brazilian Portuguese language, and then were fine-tuned and specialised for the legal sector using 210,000 legal proceedings. Vector representations of each legal document were calculated based on their embeddings, which were used to cluster the lawsuits, calculating the quality of each model based on the cosine of the distance between the elements of the group to its centroid. We noticed that models based on transformers presented better performance when compared to previous traditional NLP techniques, with the RoBERTa model specialised for the Brazilian Portuguese language presenting the best results. This methodology can be also applied to other case studies for different languages, making it possible to advance in the current state of the art in the area of NLP applied to the legal sector. 
\end{abstract}


\section{Introduction}\label{sec1}
Recent advances obtained in the area of natural language processing (NLP) have encouraged researchers to carry out scientific research which present advances in the use of a specific NLP technique to transform short texts into vector representation in which the context and semantics of the words in the document are considered. In this way, recent studies have shown that machine learning algorithms are critical tools capable of solving high-complexity problems using Natural Language Processing (NLP)~\cite{bib4}. To this end, it is possible to highlight the works of~\cite{bib5,bib6,bib7,bib8,bib9,bib10,bib11}, which, taking into account the context of words, apply techniques of word-embeddings generation, a form of vector representation of names, and consequently of documents.

However, no study has been found so far that consolidates a methodology that details the use of various NLP techniques, from the most traditional to the most current ones, using robust texts and that is tested and applied in a real case. In this way, a fertile and unexplored field was found in the legal sector to validate this methodology proposed by this present work. Thus, the use of word embeddings is essential to analyse a large set of unstructured data presented in court.

The recent history of the Brazilian Justice shows relevant transformations regarding having all its procedural documents in digital format. In 2012, the Brazilian Labour Court implemented the Electronic Judicial Process (acronym in Portuguese for “Processo Judicial Eletrônico” - PJe), and since then, all new lawsuits have become completely digital, reaching 99.9\% of cases in progress on this platform in 2020~\cite{bib1}.

Knowing the limitation of human beings analysing, in an acceptable time, a large amount of data, especially when such data appear not to be correlated, it is possible to help them in the patterns' recognition context through data analysis, computational ans statistical methods. Assuming that textual data has been exponentially increasing, patterns' examination in court documents is becoming pronouncedly challenging.

To optimise the procedural progress the Brazilian legal system provides, for ways, such as the procedural economy, the principle of speed, due process in order, and the principle of the reasonable duration of a case to ensure the swift handling of judicial proceedings~\cite{bib2}. Hence, one of the major challenges of the Brazilian Justice is swiftly meeting the growing judicial demand. At present, a specialist triages the documents and distributes the lawsuits to be judged among the team members, configuring a deviation from the main activity of the specialist, which is the production of the draft decisions. This occurrence reinforce a further increase in the congestion rate (an indicator that measures the percentage of cases that remain pending solution by the end of the base-year) and to the decrease in the supply of demand index (acronym in Portuguese for “Índice de Atendimento à Demanda” - IAD - an indicator that measures the percentage of downtime of processes compared to the number of new cases)~\cite{bib1}.

Thus, using a process grouping mechanism, it is possible to assist with the allocation of work among the advisers of the office for which the process was drawn with a good rate of similarity between the documents analysed. Furthermore, it contribute to the search for case-law\footnote{A legal term meaning a set of previous judicial decisions following the same line of understanding.} for the judgement of the cases in point, guarding the principle of legal certainty. According to Gomes Canotilho~\cite{bib3}, the general principle of legal certainty aims to ensure the individual the right to trust that the legal rulings made of their issues are based upon current and valid legal norms.

 In this way, it is possible to develop, test and deploy this methodology based on deep learning for grouping judicial processes, consolidating it for the Brazilian Labour Court from the tests and validations applied.

This work aims, therefore, to use as a baseline the results discussed by the research's Oliveira and Nascimento~\cite{bib12} comparing them with the degree of similarity between the judicial documents achieved in the inferred groups through unsupervised learning, through the application of six techniques of Natural Language Processing, which are: (i) BERT (Bidirectional Encoder Representations from Transformers) trained for general purposes for Portuguese (BERT pt-BR); (ii) BERT specialised with the corpus of the Brazilian labour judiciary (BERT Jud); (iii) GPT-2 (Generative Pre-trained Transformer 2) trained for general purposes for Portuguese (GPT-2 pt-BR); (iv) GPT-2 specialised with the corpus of the Brazilian labour judiciary (GPT-2 Jud); (v) RoBERTa (Robustly optimised BERT approach) trained for general purposes for Portuguese (RoBERTa pt-BR); and (vi) RoBERTa specialised with the corpus of the Brazilian labour judiciary (RoBERTa Jud), consolidating a methodology that was tested for Brazilian labour legal documents and making it possible to use it for other fields of justice, Brazilian or international, and who knows, making it possible to apply in documents from other areas of knowledge.

Therefore, as proposed in~\cite{bib12}, the degree of similarity indicates the performance of the model and was a result of the average similarity rate of the documents groups, which was based on the cosine similarity between the elements of the group to its centroid and, comparatively, by the average cosine similarity among all the documents of the group.

To delimit the scope of this research and make a coherent comparison, the same data as in~\cite{bib12} was applied. Thus, the data set extracted contained information from the Ordinary Appeal Brought (acronym in Portuguese for “Recurso Ordinário Interposto” - ROI) of approximately 210,000 legal proceedings\footnote{https://www.doi.org/10.5281/zenodo.7686233}. The Ordinary Appeal Brought was used as a reference, as it is regularly the type of document responsible for sending the case to trial in a higher court (2nd degree), hence creating the Ordinary Appeal (acronym in Portuguese for “Recurso Ordinário” - RO). It serves as a free plea, an appropriate appeal against final and terminative judgements proclaimed at first instance, which seeks a review of the court decision drawn up by a hierarchically superior body~\cite{bib13}.

For the present work, a literature review on unsupervised machine learning algorithms applied to the legal area was performed, using NLP, and an overview of recent techniques that use Artificial Intelligence (AI) algorithms in word embeddings generation. Then, applying some methods until obtaining results, comparing them, and finally, proposing future challenges.

\section{State-of-the-Art Review}\label{sec2}

More recent research maintain that machine learning algorithms have great potential for high complexity problem-solving. These machine learning algorithms categories can be: (i) supervised; (ii) unsupervised; (iii) semi-supervised; and (iv) via reinforcement~\cite{bib14}. This research context reviewed the literature in search of the most recent productions for the period from 2017 to 2022, through the databases (i) Google Scholar; (ii) Science Direct; and (iii) IEEE Xplorer, on unsupervised machine learning algorithms or clustering applied to the legal area using NLP.

The research revealed that, so far, few productions are dealing with the subject, which proves its complexity. We highlight the research conducted by Oliveira and Nascimento~\cite{bib12} that sought to detect the degree of similarity between the judicial documents of the Brazilian Labour Court through unsupervised learning, using NLP techniques such as (i) inverse frequency of the term document frequency (TF-IDF); (ii) Word2Vec with CBoW (Continuous Bag of Words) trained for general purposes for the Portuguese language in Brazil; and (iii) Word2Vec with Skip-gram trained for general purposes for the Portuguese language in Brazil.~\cite{Song2022} made an empirical evaluation of pre-trained language models (PLMs) for legal natural language processing (NLP) in order to verify the effectiveness of the models in this domain, which used up to 57 thousand documents.

Expanding the research for the use of Natural Language Processing applied to the judicial area, a systematic review of the literature of the challenges faced by the system of trial prediction was found, which can assist lawyers, judges and civil servants to predict the rate of profit or loss, time of punishment and articles of law applicable to new cases, using the deep learning model. The researchers describe in detail the Empirical Literature on Methods of Prediction of Legal Judgment, the Conceptual Literature on Text Classification Methods and details of the transformers model~\cite{bib15}.

Therefore, we then sought to expand the research by removing the restriction for the legal area, which revealed some publications.~\cite{bib16} Discusses using a content recommendation system based on grouping, with k-means, in similar articles through the vector transformation of the content of documents with the TF-IDF~\cite{bib17}. In~\cite{bib18}, the authors performed an automatic summarisation of texts using TF-IDF and k-means to determine the sentence groups of the documents used in creating the summary. It concludes that these studies used TF-IDF as the primary technique to vectorise textual content and that k-means is the most commonly used algorithm for unsupervised machine learning. It's also highlight the research carried out by Santana, Oliveira and Nascimento~\cite{santana_oliveira_nascimento_2022} which proposed the use of a model based on Transformers for Portuguese in the generation of word embeddings of texts published in a Brazilian newspaper, limited to 510 words, for the classification of news.

We assume that choosing the best technique of generating word embeddings requires research, experimentation and comparison of models. Many recent studies prove the feasibility of using word embeddings to improve the quality of the results of AI algorithms for pattern detection and classification, among others. However, most of the searches found use a reduced number of documents and, in addition, limit the content of these documents to a maximum of 510 words.

Mikolov et al. proposed in 2013 Word2Vec Skip-gram and CBoW, two new architectures to calculate vector representations of words considered, at the time, reference in the subject~\cite{bib6}. Then, Embeddings from Language Models (Elmo)~\cite{bib19}, Flair~\cite{bib20} and context2vec~\cite{bib21}, libraries based on the Long Short Term Memory Network (LSTM)~\cite{bib22} created distinct word embeddings for each occurrence of the word, context-aware, which allowed the capture of the meaning of the word. The LSTM models were used widely for speech recognition, language modelling, sentiment analysis and text prediction, and, unlike Recurrent Neural Network (RNN), have the ability to forget, remember and update information, thus taking a step ahead of the RNNs~\cite{bib23}.

As of 2018, new techniques for generating word embeddings emerged, with emphasis on (i) Bidirectional Encoder Representations from Transformers (BERT)~\cite{bib9}, a context-sensitive model with architecture based on a Transformers model~\cite{bib24}; (ii) Sentence BERT (SBERT)~\cite{bib25}, a “Siamese” BERT model proposed to improve BERT's performance when seeking to obtain the similarity of sentences; (iii) Text-to-Text Transfer Transformer (T5)~\cite{bib26}, a framework for treating NLP issues as a text-to-text problem, i.e. template input as text and template output as text; (iv) Generative Pre-Training Transformer 2 (GPT-2), a Transformers-based model with 1.5 billion parameters~\cite{bib10}; and (v) Robustly optimised BERT approach (RoBERTa), a model based on the BERT model, which was trained longer and used a higher amount of data~\cite{bib11}.

With this analysis, it was possible to advance in the current state of the art of NLP applied to the legal sector. By conducting a comparative study and implementation of Transformers techniques (BERT, GPT-2 and RoBERTa), using models for generic purpose in Brazilian Portuguese (pt-BR) and specialised models in the labour judiciary, to carry out the grouping of labour legal processes in Brazil using the k-means algorithm and cosine similarity. In addition to advancing in the consolidation of a methodology, validated for the Brazilian labour legal sector, which can be used in every field of justice, Brazilian and International, and in other areas of knowledge.

\section{Methodology}\label{sec3}

In this section, the protocol necessary to reproduce the results achieved and to analyse them comparatively is presented. For the implementation of the routines used in this study, we used the Python programming language (version 3.6.9) and the same libraries used in the study by Oliveira and Nascimento~\cite{bib12}.

The processing flow (pipeline) was composed of the phases: (i) data extraction; (ii) data cleaning; (iii) generation of word embeddings templates; (iv) calculation of the vector representation of the document; (v) unsupervised learning; and (vi) calculation of the similarity measure, of which phases (iii), (iv) are detailed in the follow sections, and the others phases are summarised below and for more details please refer to the work of Oliveira and Nascimento~\cite{bib12}.

\begin{itemize}
    \item data extraction: a dataset containing information from documents of the Ordinary Appeal Interposed (acronym in Portuguese for “Recurso Ordinário Interposto” — ROI) type was extracted from approximately 210,000 legal proceedings;
    \item data cleaning: was realised two two forms of preprocessing: (i) detection of the subjects of the Unified Procedural Table\footnote{Labour Justice Unified Procedural Table. Available at: https://www.tst.jus.br/web/corregedoria/tabelas-processuais} (acronym in Portuguese for “Tabela Processual Unificada”—TPU) contained in the extracted documents; and (ii) cleaning the contents of the documents, using a regular expression, for examples, remove the tags HTML, replace the name if the individuals linked to the legal cases by the "tag" "parteprocesso" (part in the process), replace the judging bodies (e.g., “Tribunal Regional do Trabalho” [Regional Labour Court]) by the “orgaojulgador” (organjudge), etc;
    \item unsupervised learning: the technique adopted was the k-means algorithm~\cite{bib17};
    \item calculation of the similarity measure: the cosine similarity measure was adopted as tool for the measurement of the quality of inferred groups.
\end{itemize}

\subsection{Generation of word embeddings templates}\label{subsec3_3}

The usage of vector representation of words, whose numerical values indicate some relationship between words in the text, is an essential technique in the machine learning problem-solving process when the data used by the model is textual.

Thus, in this research, word embeddings generated and shared for the Portuguese language were used, such as (i) BERT (large) model generated based on brWaC corpus~\cite{bib27}, composed of 2 billion and 700 thousand tokens, and published in the article BERTimbau: Pretrained BERT Models for Brazilian Portuguese~\cite{bib28}; (ii) GPT-2 (Small) model generated based on texts extracted from Wikipedia in Portuguese, and published in article GPorTuguese-2 (Portuguese GPT-2 small): a Language Model for Portuguese text generation (and more NLP tasks...)~\cite{bib29}; and (iii) RoBERTa (Base) model generated based on texts extracted from Wikipedia in Portuguese, entitled roberta-PT-BR and published in Hugging Face~\cite{bib30}.

In addition to these pre-trained models in the Portuguese language, the most recent literature suggests that using embeddings adherent to the context of the problem proposed to be solved may bring a better result. Thus, using the 210,000 documents extracted, two embedding generation techniques were applied, namely, (i) specialisation of the BERTimbau model; (ii) specialisation of the GPorTuguese-2 model; and (iii) specialisation of the roberta-pt-br model, which will be detailed below.

\subsubsection{Specialisation of Transformers models}\label{subsec3_3_1}

Recent studies show the benefits of applying for learning transfer on generalist models, which, in recent years, has significantly improved the results, reaching the state-of-the-art in NLP~\cite{bib31}. For the specialisation of Transformers models, in addition to cleaning the data, it is also necessary to adjust the data to make the most of its benefits. Of the adjustments made, two deserve highlights: (i) definition of the sentence slot; and (ii) definition of the strategy of “disguising” or masking (MASK) of the sentences’ tokens, which are detailed below.

\begin{figure}[!ht]%
\centering
\includegraphics[width=0.5\textwidth]{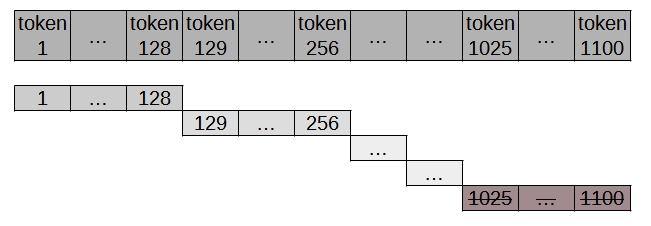}
\caption{Slot N - generation of “sentences” with 128 tokens.}\label{fig_slotN}
\end{figure}

Defining the sentence slot is a fundamental step to enable the usage of specialised data in the learning transfer from a pre-trained model. Therefore, inspired by the strategy proposed in the article Transformers: State-of-the-Art Natural Language Processing~\cite{bib32} that, for each batch of 1,000 documents, as presented in Figure~\ref{fig_slotN}, all content is concatenated and sentences of 128 tokens created, if the last “sentence” of this lot is less than 128 tokens this “sentence” is disregarded, other detailed approaches have been tested later.

\begin{figure}[!ht]%
\centering
\includegraphics[width=0.5\textwidth]{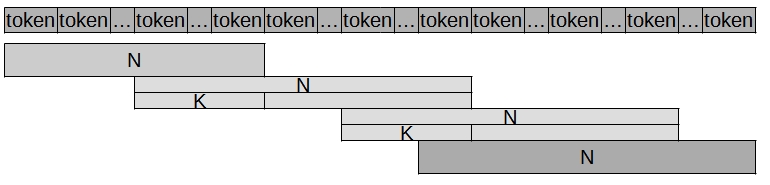}
\caption{N/K slot.}\label{fig_slotN_K}
\end{figure}

In order to reduce the loss of context of words at the edges of the sentence, the proposed approach, entitled Slot N/K, generated “sentences” with N tokens from the concatenation of 1,000 documents, as detailed below and illustrated in Figure~\ref{fig_slotN_K}.

\begin{itemize}
\item Initial Slot: “sentence” formed by the first N tokens;
\item Intermediate slots: “sentence” formed by N tokens counted from the N-K token of the previous “sentence”, where K is the number of return tokens;
\item Final Slot: “sentence” formed by the last N tokens.
\end{itemize}

\begin{figure}[!ht]%
\centering
\includegraphics[width=0.5\textwidth]{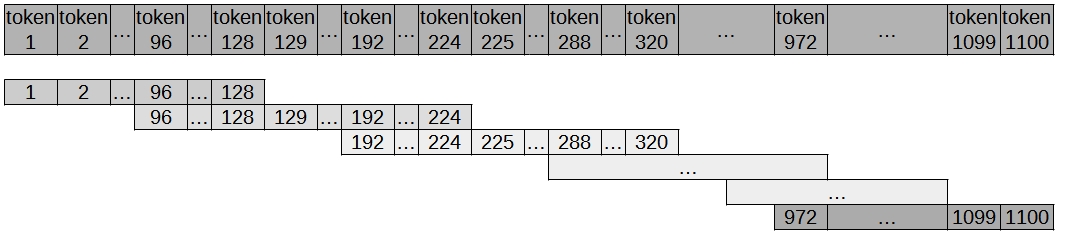}
\caption{Slot 128/32.}\label{fig_slot1238_32}
\end{figure}

From the above-detailed approach, simulations performed with the settings (i) Slot 128/16; (ii) Slot 128/32; (iii) Slot 128/64; (iv) Slot 256/64; (v) Slot 512/64; and (vi) Slot 64/16, comparing them with each other and with the approach proposed by~\cite{bib32}. The Slot 128/32 approach was selected for achieving the best performance in the specialisation of the Transformers model in Portuguese with the corpus of the judiciary (Figure~\ref{fig_slot1238_32}).

For learning transfer, depending on the Transformers model used, a token masking strategy for each sentence is applied, using Masked Language Models (MLM) for BERT and Causal Language Models (CLM) models for GPT-2 models. While the CLM is trained unidirectionally in order to predict the next word based on the preceding words~\cite{bib33}, the MLM has a two-way approach to predict the masked words of the sentence.

Hence, for the transfer of learning of BERT models, inspired by the article Transformers: State-of-the-Art Natural Language Processing~\cite{bib32} that used the masking rate of 15\%, simulations were performed using the masking rate of 15\% and the masking rate of 25\%, reaching, with the rate of 15\%, the best result in the specialisation of the BERT model in Portuguese with the corpus of the judiciary.

\subsection{Calculation of the vector representation of the document}\label{subsec3_4}

Vector representation techniques of words (word embeddings) such as (i) BERT; (ii) GPT-2; and (iii) RoBERTa need to undergo a transformation in order to, from the word embeddings, calculate the vector representation of the document (document embeddings).

It is initially necessary to detail how to obtain word embeddings for Transformers techniques. One of the advantages of Transformers techniques over previous word embeddings techniques, such as Word2Vec, is the ability to capture the vector representation of the word according to the global context, meaning that the same word can have more than one vector representation. It becomes more evident when highlighting the word “bank” ($banco$ in Pt-BR) in the following two sentences (i) I go to the bank ($banco$ in Pt-BR) to withdraw money; and (ii) I will sit on the bench ($banco$ in Pt-BR) of the square; where, with Word2Vec, the vector representation of the word “bank” is unique regardless of the phrase and with BERT, GPT-2 and RoBERTa word embeddings are different.

Therefore, for Transformers templates, it is necessary to “divide” the entire document into “slots” of sentences. Considering that, unlike the GPT-2 model, the BERT and RoBERTa models have a limitation of up to 512 tokens per sentence and require that the first and last tokens be special, respectively [CLS] and [SEP], the slot size has been set at 510 tokens per sentence.

As indicated in the Section~\ref{sec2} (~\nameref{sec2}), unlike the present work, which has very large documents, current research limits the texts used in the vector transformation to up to 510 words. Thus, we developed strategies to obtain all the word embeddings of the document, whose words of the generated sentences kept the context according to the complete file. These approaches consist, similar to that presented in Figure~\ref{fig_slotN_K}, in bringing about sentences with 510 tokens as detailed below:

\begin{itemize}
\item Initial Sentence: “sentence” formed by the first 510 tokens;
\item Intermediate sentences: “sentence” consisting of 510 tokens counted from token N - K of the previous “sentence”, where K was set empirically to value 64;
\item Final Sentence: “sentence” formed by the last 510 tokens;
\end{itemize}

Therefore, the sentences generated from each document have coincident tokens chosen to ensure greater adherence to the token context in the file. To this end, we tested two different approaches: (i) averages of word embeddings of coincident tokens; and (ii) use of the first 32 coincident tokens of the previous sentence and the last 32 coincident tokens of the current sentence, which showed better results in the simulations performed.

\begin{figure}[!ht]%
\centering
\includegraphics[width=0.5\textwidth]{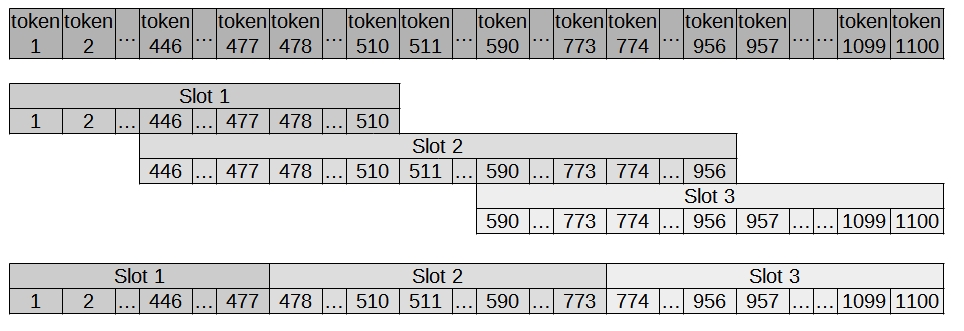}
\caption{Word embeddings generation strategy.}\label{fig_doc_embeddings}
\end{figure}

Hence, as shown in Figure~\ref{fig_doc_embeddings}, the return tokens that are coincident between the current and previous sentences are used as follows: (i) the first 32 coincident tokens of the previous sentence (for example, tokens 446 to 477 from Slot 1 exemplified in Figure~\ref{fig_doc_embeddings}); and (ii) the last 32 coincident tokens of the sentence in question (for example, Tokens 478 to 510 from Slot 2 exemplified in Figure~\ref{fig_doc_embeddings}). It is worth noting that the last sentence slot must contain 510 tokens, as well as the others, and coincident tokens tapped as follows: (i) the first half of the coincident tokens of the previous sentence (for example, tokens 590 to 773 from Slot 2 exemplified in Figure~\ref{fig_doc_embeddings}); and (ii) the second half of the coincident tokens of the sentence in question (for example, tokens 774 to 956 from Slot 3 exemplified in Figure~\ref{fig_doc_embeddings}).

After obtaining the word embeddings, the same technique used in the research by Oliveira and Nascimento was chosen to generate the document embeddings, that is, the average of the word embeddings of the words in the document, weighting them with the TF-IDF.

Consequently, to enable an overview, Table~\ref{tab_parameters} summarises the parameters used for training the six models used in this research.

\begin{table}[!ht]
\caption{Parameters used for training the six models.\label{tab_parameters}}
\centering
\setlength{\tabcolsep}{3.7pt}
\begin{tabular}{lrccccc}
\multicolumn{2}{r}{\makecell[r]{BERT\\imbau}} & \makecell{BERT\\Jud.} & \makecell{GPortu\\guese-2} & \makecell{GPT-2\\Jud.} & \makecell{roberta\\-pt-br} & \makecell{RoBERTa\\ Jud.} \\
\hline
\makecell[l]{Data for\\traning} & \makecell{brWac\\corpus} & \makecell{210 K\\ROIs} & \makecell{Wikipe-\\dia in\\Portu-\\guese} & \makecell{210 K\\ROIs} & \makecell{Wikipe-\\dia in\\Portu-\\guese} & \makecell{210 K\\ROIs} \\
\hline
\makecell[l]{Tokenization\\type} & \multicolumn{2}{c}{word-piece} & \multicolumn{2}{c}{\makecell{byte-level\\BPE}} & \multicolumn{2}{c}{\makecell{byte-level\\BPE}} \\
\hline
\makecell[l]{Model\\details} & \multicolumn{2}{c}{\makecell{24-layer,\\1024-hidden,\\16-heads, \\340M parameters}} & \multicolumn{2}{c}{\makecell{12-layer,\\768-hidden,\\12-heads, \\117M parameters}} & \multicolumn{2}{c}{\makecell{12-layer,\\768-hidden,\\12-heads, \\125M parameters}} \\
\hline
\makecell[l]{Token\\mask type} & \multicolumn{2}{c}{Masked} & \multicolumn{2}{c}{Masked} & \multicolumn{2}{c}{Causal} \\
\hline
\end{tabular}
\end{table}

Moreover, after going through the stages of generating the unsupervised machine learning model and calculating the similarity measure, as defined by Oliveira and Nascimento~\cite{bib12} a two-dimensional graphic representation of the vector representation of the documents was generated, using the T-Distributed Stochastic Neighbor Embedding (t-SNE) reduction technique, which minimizes the divergence between two distributions by measuring the similarities between pairs of input objects and the similarities between pairs of corresponding low-dimensional points in the embedding~\cite {bib34}.

\section{Results and Discussions}\label{sec4}

Applying the methodology as previously detailed, this research shows how natural language processing techniques in conjunction with machine learning algorithms are paramount in optimising the operational costs of the judicial process, such as the aid of document screening and procedural distribution. It grants working time optimisation since it allows the experts time to be devoted to their core activity.

\begin{figure}[!ht]%
\centering
\includegraphics[width=0.5\textwidth]{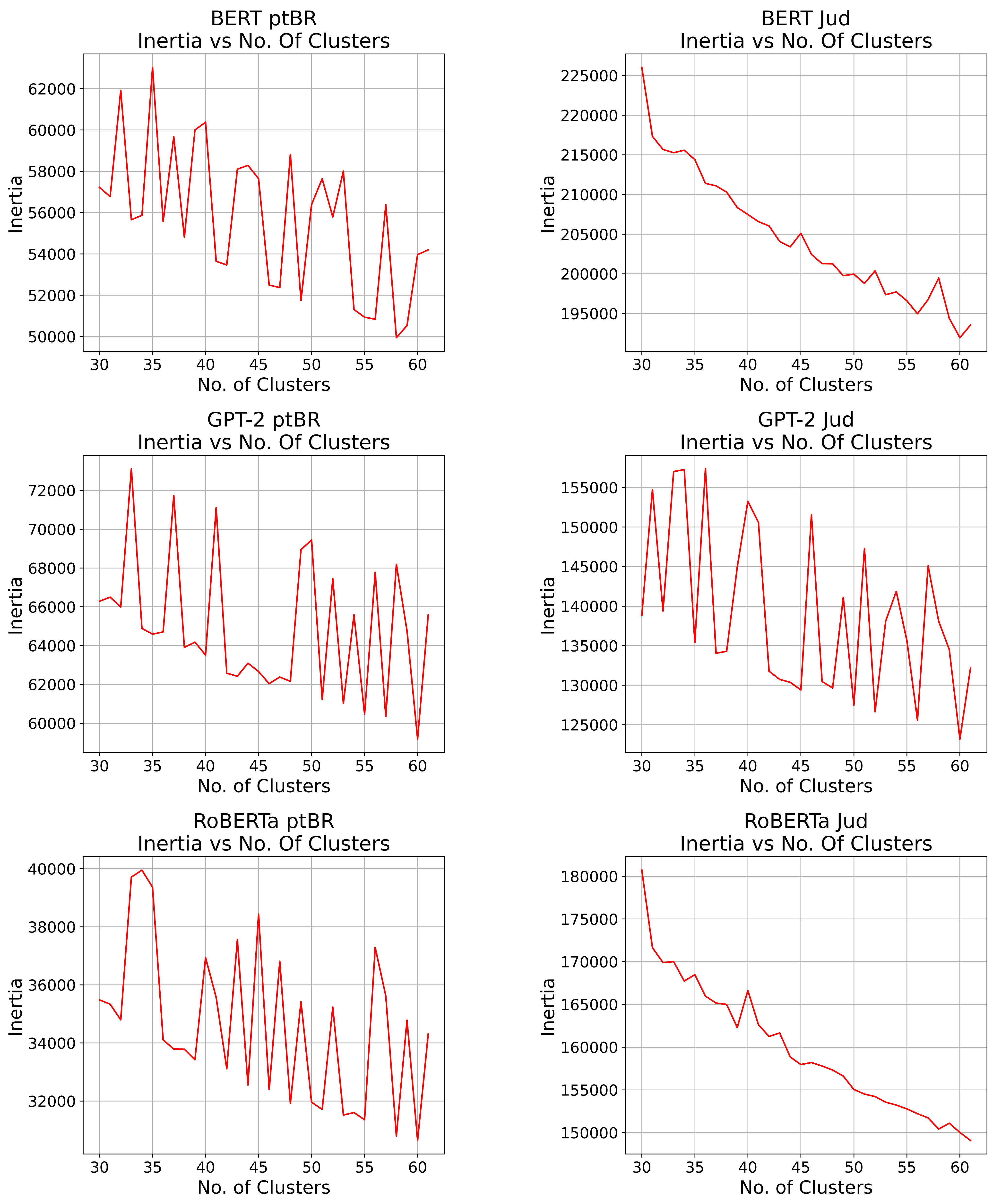}
\caption{Inertia charts constructed by using the elbow method for determining the best number of clusters for each approach.}\label{fig_inertia}
\end{figure}

In order to use the unsupervised learning algorithm, k-means, it was necessary to define the ideal K to offer to the clustering algorithm, we used the inertia calculation, which measures how well the data set grouped through k-means. The inertia calculation is the sum of the square of the Euclidean distance from each point to its centroid, seeking to reach the K with the lowest inertia. The tendency is that the higher the K, the lower the inertia, then we used the elbow method to find the K where the reduction of inertia begins to decrease. Then, we used 31 values for K within the range from 30 to 61, with intervals per unit, selecting according to the elbow technique the K that generated the best grouping, as showed in Figure~\ref{fig_inertia}.

From obtaining the best K, the k-means template was trained and, with the grouping performed by the model, we calculated (i) the average similarity between the documents of each group, thus allowing an overview of the distribution of documents in the groups generated by each NLP technique; and (ii) the mean similarity between the group's documents and their centroid, making possible to indicate which technique achieved the best performance.

To demonstrate the progress brought by this research, Table~\ref{tab_results_base_line} presents the results extracted from the study~\cite{bib12}, which established a baseline for research on the use of NLP techniques applied to the legal environment for the same purpose. We highlight the Word2Vec Skip-gram pt-BR technique, which presented itself, in that research, as the best option for generating word embeddings aiming to group judicial documents of the Ordinary Appeal Brought type.

\begin{table}[!ht]
\caption{Statistical data extracted from the work “Clustering by Similarity of Brazilian Legal Documents Using Natural Language Processing Approaches”~\cite{bib12}\label{tab_results_base_line}}
\centering
\setlength{\tabcolsep}{3.7pt}
\begin{tabular}{lcccccccc}
\multicolumn{1}{c}{Type} & Groups & Mean & Std. & Min. & 25\% & 50\% & 75\% & Max. \\
\hline
TF-IDF & 49 & 0.624 & 0.172 & 0.247 & 0.502 & 0.586 & 0.164 & 0.964 \\
\hline
\makecell[l]{Word2Vec\\CBow\\ptBR} & 59 & 0.947 & 0.063 & 0.764 & 0.935 & 0.979 & 0.991 & 0.999 \\
\hline
\makecell[l]{Word2Vec\\Skip-gram\\ptBR} & 34 & 0.948 & 0.061 & 0.796 & 0.925 & 0.976 & 0.992 & 0.999 \\
\hline
\end{tabular}
\end{table}

\begin{figure}[!ht]%
\centering
\includegraphics[width=0.5\textwidth]{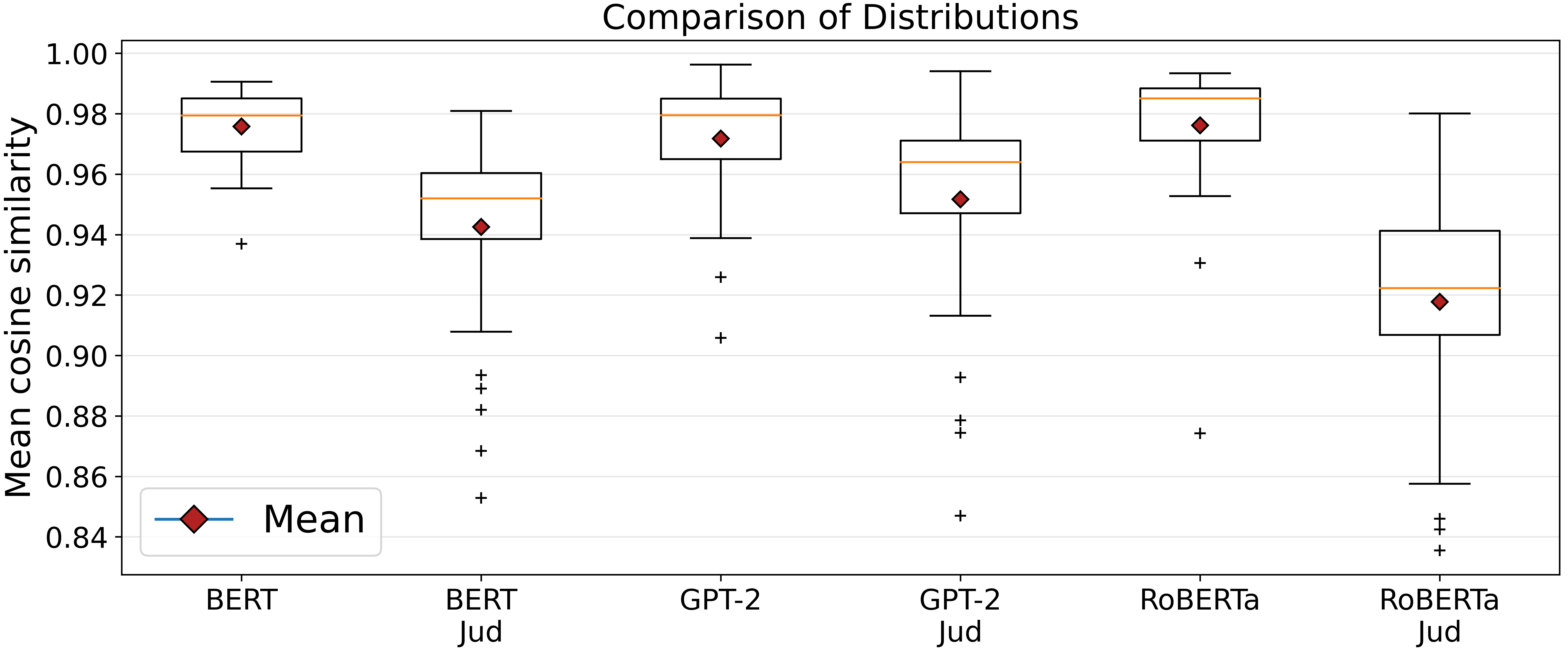}
\caption{Comparison chart of the distribution of the average similarity between the group documents. The more cohesive the boxes and the fewer outliers, the better.}\label{fig_boxplot_group}
\end{figure}

\begin{figure}[!ht]%
\centering
\includegraphics[width=0.5\textwidth]{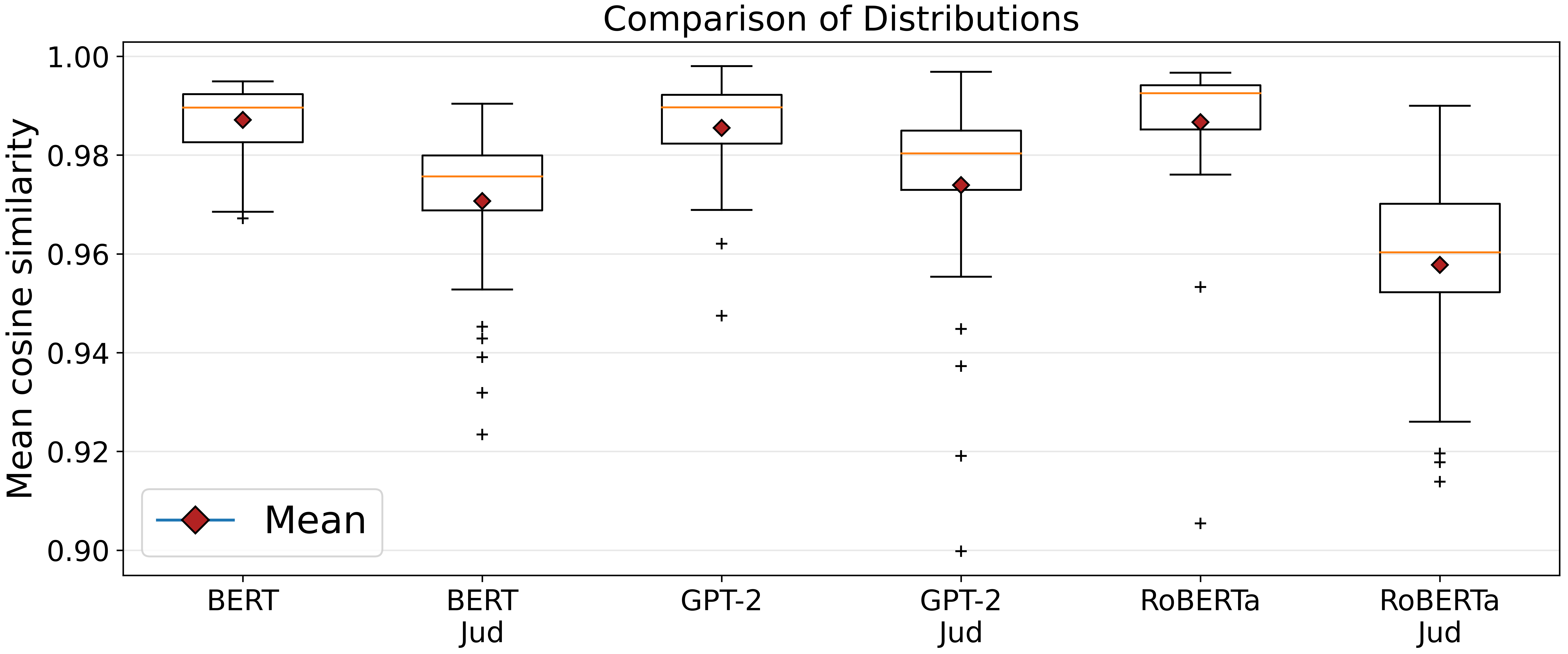}
\caption{Comparison chart of the distribution of the average similarity of the group documents to their centroid. The more cohesive the boxes and the fewer outliers, the better.}\label{fig_boxplot}
\end{figure}

Consequently, the statistical data of the average similarity between the documents of each group and the average similarity of the group documents for their centroid presented respectively in Table~\ref{tab_results_all_elements} and Table~\ref{tab_results_elements_centroid}, highlighted in bold for the best result value of each metric and projected in the comparative distribution chart (Figure~\ref{fig_boxplot_group} and Figure~\ref{fig_boxplot}), show that the generalist word embeddings in Portuguese (pt-BR) achieved superior results when compared to the specialised legal corpus word embeddings. The proximity of the results among the generalist models is also noteworthy. However, for the expert model, this proximity was observed only between the BERT Jud models and GPT-2 Jud.

\begin{table}[!ht]
\caption{Statistics of the cosine similarity between all elements of the group, where the pt-BR models are generalist and the Jud. models are specialised. The best results are highlighted in bold.}\label{tab_results_all_elements}
\centering
\setlength{\tabcolsep}{3.7pt}
\begin{tabular}{lcccccccc}
\hline
\multicolumn{1}{c}{\makecell{Transformer\\Model}} & Groups & Mean & Std. & Min. & 25\% & 50\% & 75\% & Max. \\
\hline
BERT ptBR & 35 & 0.976 & \textbf{0.012} & \textbf{0.937} & 0.967 & 0.979 & 0.985 & 0.991 \\
BERT Jud & 36 & 0.943 & 0.031 & 0.853 & 0.938 & 0.952 & 0.960 & 0.981 \\
GPT-2 ptBR & \textbf{33} & 0.972 & 0.020 & 0.906 & 0.965 & 0.979 & 0.985 & \textbf{0.996} \\
GPT-2 Jud & 36 & 0.952 & 0.034 & 0.847 & 0.947 & 0.964 & 0.971 & 0.994 \\
\makecell[l]{RoBERTa\\ptBR} & 34 & \textbf{0,976} & 0,023 & 0,874 & \textbf{0,971} & \textbf{0,985} & \textbf{0,988} & 0,993 \\
\makecell[l]{RoBERTa\\Jud} & 39 & 0,918 & 0,035 & 0,835 & 0,927 & 0,922 & 0,941 & 0,980 \\
\hline
\end{tabular}
\end{table}

\begin{table}[!ht]
\caption{Statistics of the cosine similarity of the group elements to the centroids, where the pt-BR models are generalist and the Jud. models are specialised. The best results are highlighted in bold.}\label{tab_results_elements_centroid}
\centering
\setlength{\tabcolsep}{3.5pt}
\begin{tabular}{lcccccccc}
\hline
\multicolumn{1}{c}{\makecell{Transformer\\Model}} & Groups & Mean & Std. & Min. & 25\% & 50\% & 75\% & Max. \\
\hline
BERT ptBR & 35 & \textbf{0.987} & \textbf{0.007} & \textbf{0.967} & 0.983 & 0.970 & 0.992 & 0.995 \\
BERT Jud & 36 & 0.971 & 0.016 & 0.923 & 0.969 & 0.976 & 0.980 & 0.990 \\
GPT-2 ptBR & 33 & 0.985 & 0.011 & 0.947 & \textbf{0.985} & 0.990 & 0.992 & \textbf{0.998} \\
GPT-2 Jud & 36 & 0.974 & 0.021 & 0.900 & 0.973 & 0.980 & 0.985 & 0.997 \\
\makecell[l]{RoBERTa\\ptBR} & 34 & 0.987 & 0.017 & 0.905 & 0.985 & \textbf{0.992} & \textbf{0.994} & 0.997 \\
\makecell[l]{RoBERTa\\Jud} & 39 & 0.958 & 0.019 & 0.914 & 0.952 & 0.960 & 0.970 & 0.990 \\
\hline
\end{tabular}
\end{table}

When comparing the values presented in Table~\ref{tab_results_all_elements} and Table~\ref{tab_results_elements_centroid}, it is noteworthy that the results in Table~\ref{tab_results_all_elements} are slightly lower in all cases. From this, it is noticeable that the measurement of similarity as in Table~\ref{tab_results_all_elements} might reduce the similarity rate since there may be elements positioned altogether opposite in the group. From Figure~\ref{fig_boxplot_group} and Figure~\ref{fig_boxplot}, it is also possible to verify that the groupings generated by all techniques are very cohesive, especially in the generalist techniques cases, which created fewer groupings in the range of outliers than the expert techniques.

Since most of the techniques achieved results close to each other, we considered it important to present the time spent for the processing of each Transformer technique, with the use of a computer with 40 physical nuclei and 196 GB of memory, in the generation of numerical representation of approximately 210,000 judicial documents of the Ordinary Appeal Brought type. As presented in Table~\ref{tab_peformance}, GPT-2 reached an average vectorisation of documents per minute much higher than BERT. However, as expected, RoBERta further outperformed BERT and GPT-2, as~\cite{bib11} performance can be improved when trained for more extended periods, with larger batches, over more data, without using the prediction of the next sentence strategy, in addition to training longer sequences with dynamically changed standard masking. In this way, the performance evaluation is fundamental when it comes to documents in large quantity and whose content has many words, bringing more prominence to this research because of this particularity.

\begin{table}[!ht]
\caption{Average processed documents per minute for each model highlighted in bold for the best result value.}\label{tab_peformance}
\centering
\begin{tabular}{lc}
\hline
\multicolumn{1}{c}{Transformer Model} & \makecell{Average number of documents\\processed per minute} \\
\hline
BERT ptBR & 6.45 \\
BERT Jud & 9.62 \\
GPT-2 ptBR & 29.40 \\
GPT-2 Jud & 29.03 \\
RoBERTa ptBR & \textbf{55.31} \\
RoBERTa Jud & 53.73 \\
\hline
\end{tabular}
\end{table}

Given the above, among all the techniques evaluated, the RoBERTa pt-BR technique was the best option for generating word embeddings for judicial documents clustering of the Ordinary Appeal Brought type. Although the BERT pt-BR technique achieved a slightly better result (a difference minor than 0.01), it was computationally inefficient in document processing gpt-2 pt-BR and RoBERTa pt-BR techniques. 

It is important to stress that the results of this research (Table~\ref{tab_results_elements_centroid}) showed relevant advances in contrast to the results presented in the previous research (Table~\ref{tab_results_base_line}), in which the best average cosine similarity of the elements of the group to the centroid was, respectively, 0.98 and 0.94. So, as a consolidation of the methodology, it can be seen that with the advance of approximately 4 points with the use of the Transformers architecture when comparing the results presented by the research by Oliveira and Nascimento~\cite{bib12} and the present work, it may indicate that the sequence of techniques presented by this methodology provided subsidies for its use in other areas of Brazilian and international justice, and could also be used in other areas of knowledge.

A fact to be analysed in the results presented is that specialised word embeddings techniques showed slightly worse results. Its occurrence is due to the general techniques in Portuguese being trained with a much larger corpus than the one used to refine the generalist model. This fact is also reported by Ruder et al.~\cite{bib31}, featuring a behaviour similar to that found in the present study, in which the corpus of the base model is much larger than the specialised corpus used.

\begin{figure}[!ht]%
\centering
\includegraphics[width=0.75\textwidth]{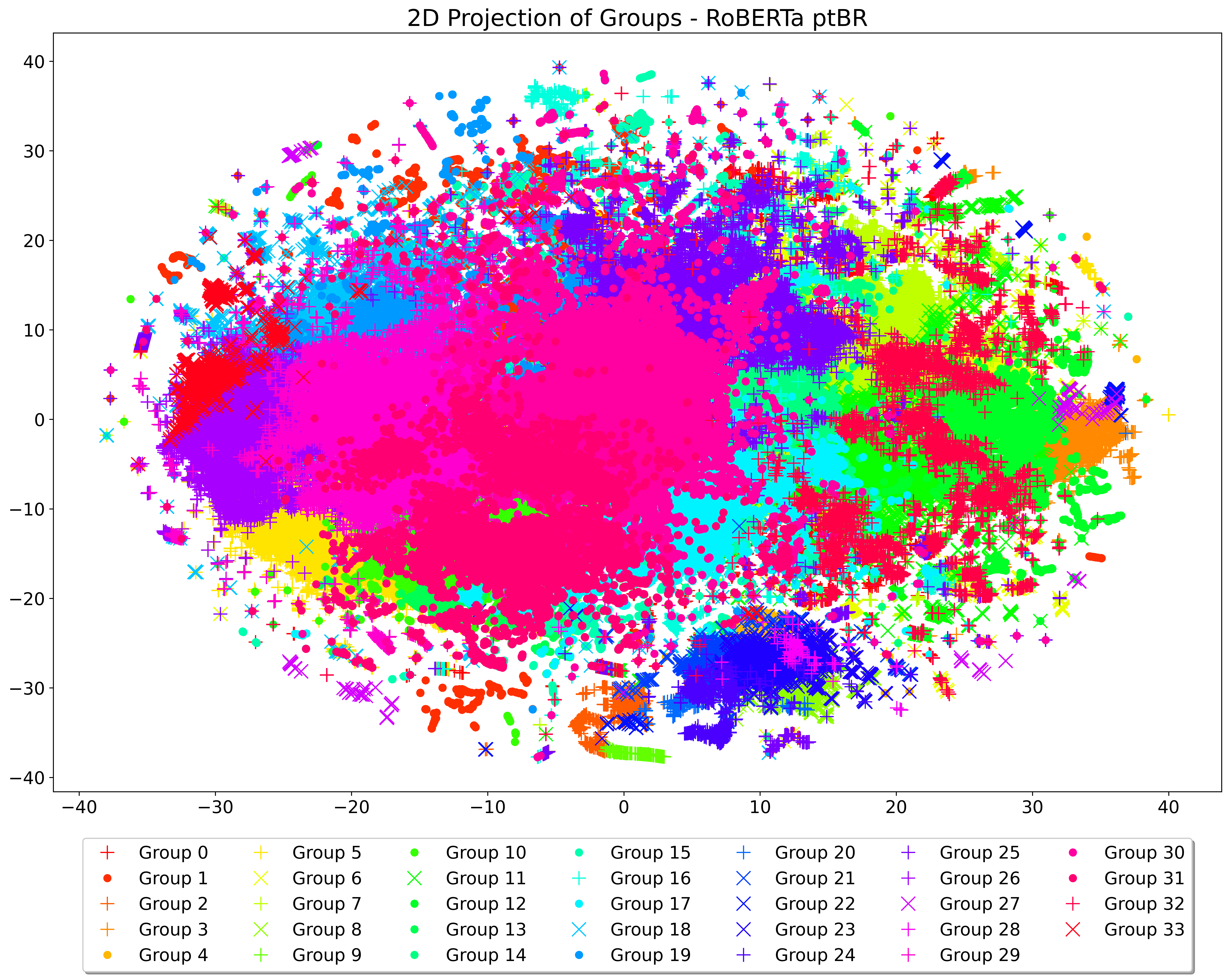}
\caption{Groups of documents formed by the RoBERTa pt-BR technique projected in two dimensions using the test dataset.}\label{fig_2D_RoBERTa}
\end{figure}

\begin{figure}[!ht]%
\centering
\includegraphics[width=0.75\textwidth]{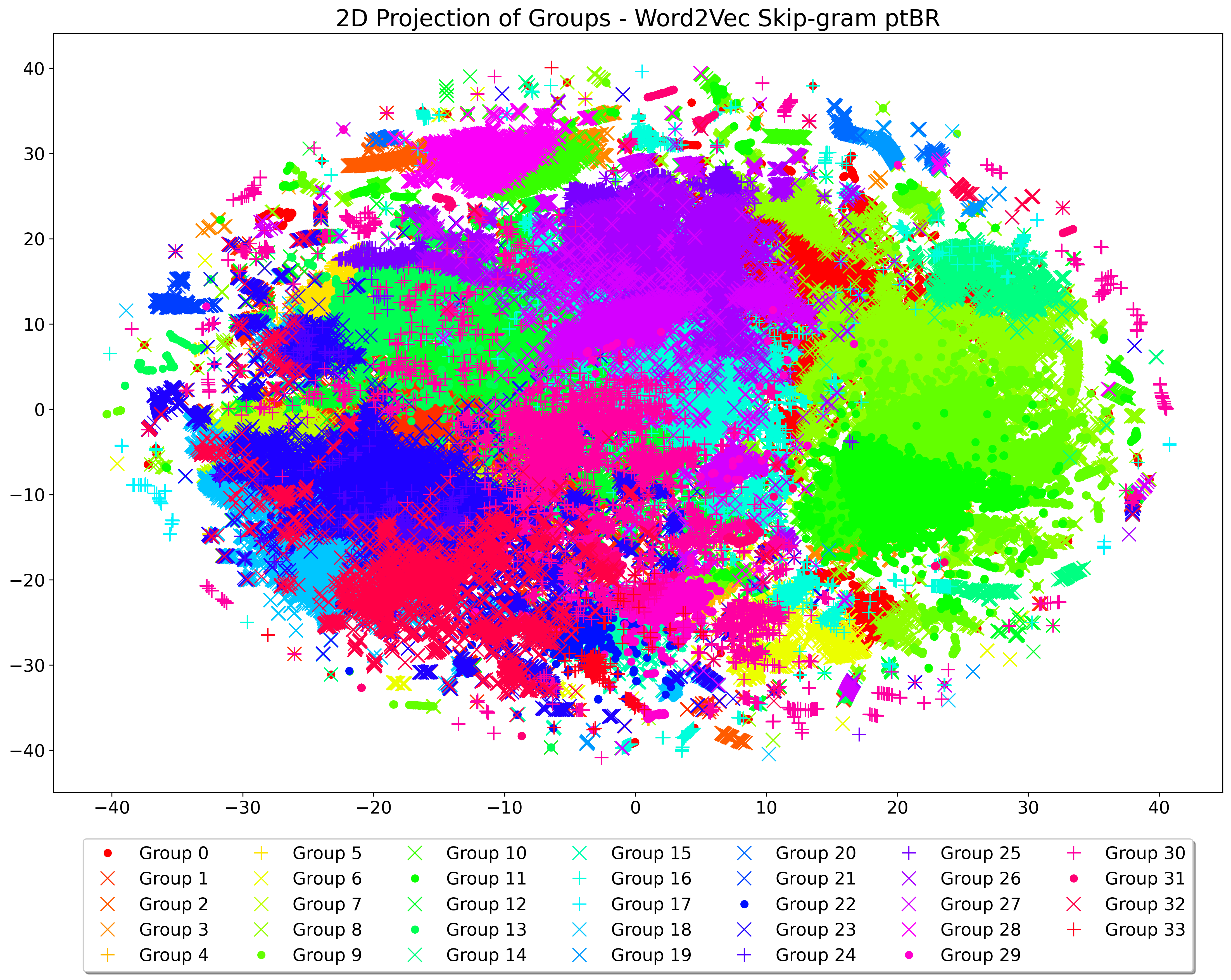}
\caption{Groups of documents formed by the Word2Vec Skip-gram technique projected in two dimensions using the test dataset~\cite{bib12}.}\label{fig_2D_W2V_SKIP}
\end{figure}

The results achieved by each approach of the entire methodology developed can be visualised in a two-dimensional projection of the groups formed in the nine techniques (i) TF-IDF; (ii) Word2Vec CBoW ptBR; (iii) Word2Vec Skip-gram ptBR; (iv) BERT ptBR; (v) BERT Jud.; (vi) GPT-2 ptBR; (vii) GPT-2 Jud.; (viii) RoBERTa ptBR; e (ix) RoBERTa Jud., which are made available as supplemental material for comparison purposes. Thus, after a qualitative analysis, it is evident in the images that the groups formed from the RoBERTa pt-BR (Figure~\ref{fig_2D_RoBERTa}) are much better defined, which corroborates the findings previously explained in this study. Furthermore, it is important to highlight that when we visualise the two-dimensional projection of the best technique presented by Oliveira and Nascimento~\cite{bib12} (Figure~\ref{fig_2D_W2V_SKIP}) and in the present work (Figure~\ref{fig_2D_RoBERTa}) solidifies the importance the consolidation of a methodology for pattern detection using natural language processing applied to thousands of documents with very large contents.

\section{Conclusions and Future Works}\label{sec5}

Applying AI techniques as a tool for pattern detection in legal documents has been proven as a viable and effective solution, showing satisfactory results that can underpin the practice of legal work, even more in certain circumstances where technicians and analysts are overwhelmed by huge volumes of work. In this way, it was possible to develop, test and deploy this methodology based on deep learning for grouping judicial processes, applying it, as a case study, for the Brazilian Labour Court, enabling the use of this methodology in other languages for the legal sector, and potentially for other areas of research, as well.

Results showed this methodology to be very promising, due to the noticeable improvement in the Average Similarity Rate in the groups formed from the use of all NLP techniques applied in this work for clustering legal documents through unsupervised machine learning. In addition, this methodology dealt with documents composed of a lot of contents, which brought a difference to what has been seen in the scientific literature so far. Of all the techniques evaluated, the RoBERTa pt-BR technique stands out as the best option for the generation of vector representations of documents based on  the embeddings for the task of clustering legal documents of the Ordinary Appeal Brought type, highlighting that the use of the best NLP technique to obtain word embeddings for finally generating the document embedding ensured a considerable improvement in the groupings. BERT pt-BR technique also presented interested results since its quantitative rates were slightly better than RoBERTa pt-BR, even though it did not reach an execution time as satisfactory as RoBERTa pt-BR. Hence, as detailed in the methodology, this performance characteristic of the RoBERTa models' architecture is key when dealing with thousands of documents with very large contents, which becomes the case when processing legal documents in courts.

On the other hand, the specialised models with the corpus of the judiciary, in general, did not achieve better results than the generalist one. Despite of this, we believe that the specialisation of BERT, GPT-2 and RoBERTa with a more robust legal corpus could achieve even better results, in such a way that the creation of generalist models for a given language for its legal area, that is, second-level foundation NLP models for the target language with a focus on the legal sector, would allow the creation and specialisation of new and more robust NLP models focused on the diverse legal areas, leveraging the results achieved since the language used in the legal environment has its own characteristics.

Furthermore, based on the methodology developed and evaluated in this work, a tool called GEMINI was developed for the Brazilian Labour Court, which allowed to assist in the search for jurisprudence, in the distribution of work among advisers and in the detection of possible opportunities to standardise the interpretation of the legal understanding within the courts, that is, to establish Cases of Uniformity of Jurisprudence. This tool was made available for implementation in all of the twenty-four Brazilian Labour Courts and, based on the suggested groupings, helped to speed up the progress of the resolution of the processes as reported by official websites, in Portuguese, of the Brazilian Labour Justice~\cite{noticiaGEMINI1, noticiaGEMINI2, noticiaGEMINI3, noticiaGEMINI4}.

Therefore, for future work, we suggest deepening the specialisation of BERT, GPT-2 and RoBERTa for the judiciary and evaluating whether the new embeddings generated would improve the overall performance of clustering. In addition, new possibilities arise, such as validating the word embeddings generated for other types of legal documents and areas, using them in other applications, such as the generation of decision drafts and classification of documents and processes. It is also worth delving into techniques for texts transformation into their vector representations faster in their word embeddings.

\section*{Acknowledgements}
The authors thank the Reference Centre on Artificial Intelligence and the Supercomputing Centre for Industrial Innovation, both from SENAI CIMATEC, as well as the SENAI CIMATEC/NVIDIA AI Joint Lab, and the Surrey Institute for People-Centred AI at the University of Surrey, for all the scientific and technical support. The authors also thank the Regional Labour Court of the 5th Region for making datasets available to the scientific community and contributing to research and technological development.

%Bibliography
\bibliographystyle{unsrt}  
\bibliography{references}

\end{document}